# Vision-Based Tactile Sensing for the Perception of the Object's Compliance and Hardness

Muxing Huang, Zibin Chen, Weiliang Xu, Zilan Li, Yuanzhi Zhou, Guoyuan Zhou, Wenjing Chen, and Xinming Li, Member, IEEE

***Abstract*—**Object compliance perception enables the identification of soft materials, supporting tasks such as fruit detection and assisted medical palpation. Compliance perception requires sensing an object's deformation and contact forces. Existing vision-based tactile sensing for compliance perception usually depends only on force or deformation. However, deformation-based approaches cannot reliably quantify object compliance, while force-based methods are overly sensitive to geometric variations. To address these limitations, this paper presents a framework that fuses temporal force sequences and deformation field information. Specifically, time-varying contact forces are inferred from tactile image sequences via a neural network, while deformation characteristics are encoded using depth maps. Crucially, the temporal force sequence is used as a dynamic force-response feature for compliance recognition. In standard experiments, the force prediction error reaches 0.06 N within a measurement range of 12 N. The proposed method achieves a 98.0% Shore-hardness classification accuracy for samples ranging from 10 HA to 80 HA. In practical scenarios, including abnormal fruit detection and soft matter, the overall accuracy exceeds 98.5%. This method improves the compliance perception ability of artificial tactile systems and facilitates their deployment in embodied perception applications.

## I. INTRODUCTION

The perception of compliance is an active exploration process that requires applying a controllable contact force to an object while observing its deformation state as the force changes [1]. Humans judge the state of an object based on the perception of compliance by pressing and rebounding with their fingertips [2,3]. For example, people can judge the ripeness of a fruit by perceiving its compliance; doctors rely on applying specific pressure during palpation and perceiving the deformation characteristics to identify the type of mass.

Humans accurately perceive object compliance by synergistically fusing contact force and surface deformation cues [3]. Consequently, vision-based tactile sensors have emerged as a promising option for compliance perception, as they enable the simultaneous extraction of these two critical physical modalities. Recent studies [4-8] further underscore that contact deformation field data are indispensable for achieving high-precision compliance estimation. For instance, Yuan et al. [9] developed a GelSight tactile sensor to construct time-series deformation fields, employing a neural network to discriminate material compliance [10]. Because identical deformation patterns can arise from varying forces or contact conditions, relying solely on deformation makes it difficult to assess object compliance. To address this ambiguity, researchers have turned to force dynamics for quantitative analysis and identification [11-13]. For example, Jiang et al. [11] developed a neural network-based force perception model that achieves real-time quantitative estimation of object stiffness. However, for soft objects with diverse geometric features, relying solely on force signal recognition remains insufficient for accurate compliance perception.

To address the above challenges, this paper proposes a vision-based tactile perception framework that integrates temporal force sequences with deformation-field information for object compliance recognition. As shown in Fig. 1, the sensor captures tactile image sequences during indentation. A neural network is then used to infer the force evolution from the image sequence, while deformation-field features are extracted to characterize contact geometry. By jointly modeling force evolution and deformation features, the proposed method mimics the coordinated use of force and deformation cues in human tactile perception. The main contributions are summarized as follows:

(1) A compliance perception strategy is proposed by integrating temporal force sequences and deformation fields from vision-based tactile images. This strategy combines dynamic force responses with contact deformation features, improving the robustness of compliance-related perception under different object geometries.

(2) The proposed sensing system achieves a force prediction error of 0.06 N, a deformation-geometry recognition accuracy of 93.75%, and a compliance classification accuracy of 98.0%. These results demonstrate the effectiveness of fusing force evolution and deformation-field information for distinguishing objects with different geometries and Shore-hardness levels.

(3) The proposed framework further supports soft-object recognition in practical scenarios, achieving an overall recognition accuracy exceeding 98.5% in tasks such as defective fruit detection. These results highlight the potential of vision-based tactile sensing for robotic compliance perception and embodied interaction.

This work was supported in part by the South China Normal University start-up fund, and in part by Guangdong Basic and Applied Basic Research Foundation 2025A1515140161. *(Muxing Huang and Zibin Chen contributed equally to this work.) (Corresponding author: Xinming Li.)*

The authors are with the Guangdong Basic Research Center of Excellence for Structure and Fundamental Interactions of Matter, Guangdong Provincial Key Laboratory of Nanophotonic Functional Materials and Devices, School of Optoelectronic Science and Engineering, South China Normal University, Guangzhou 510006, China (corresponding author e-mail: xmli@m.scnu.edu.cn).

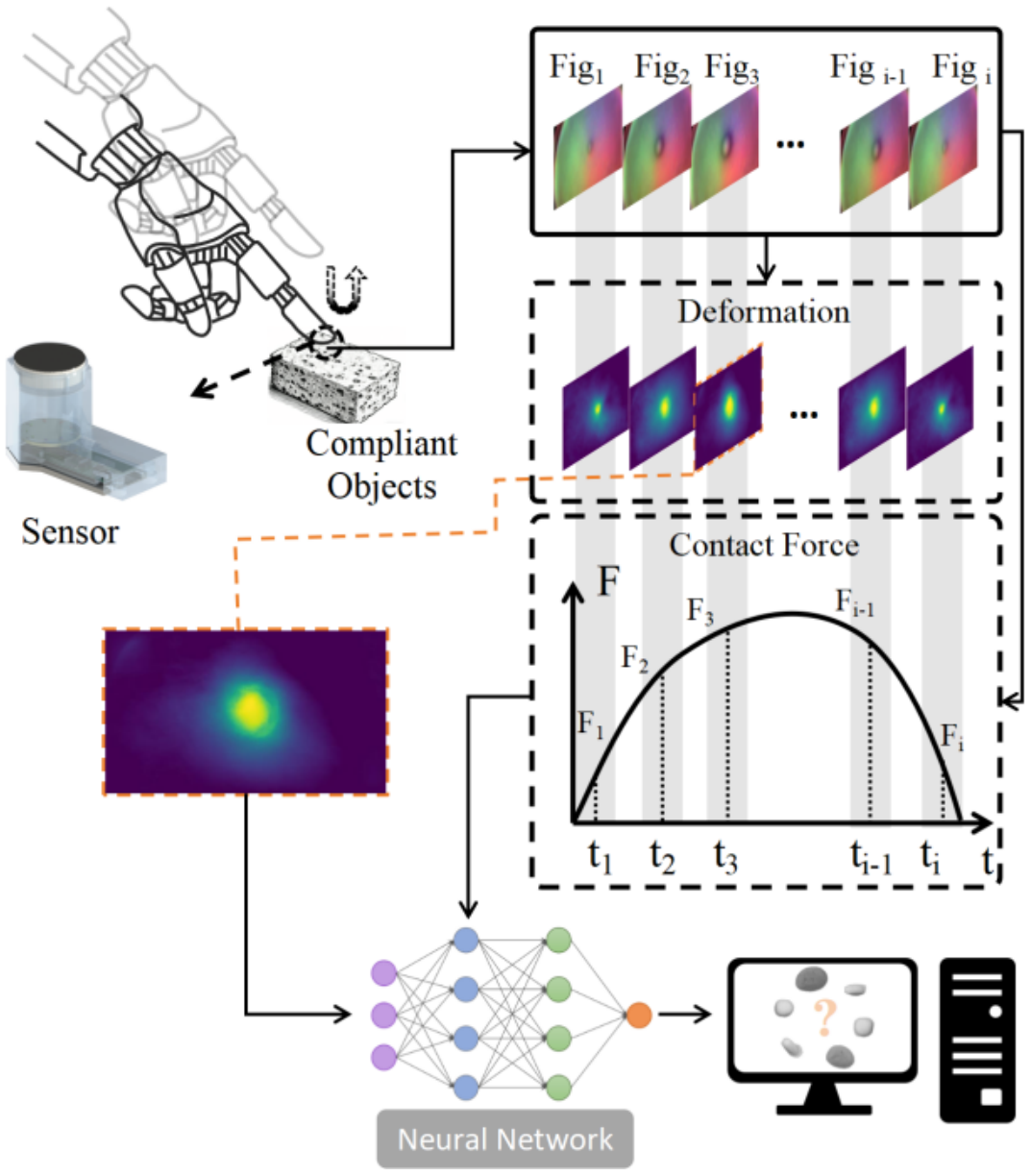

Fig. 1. The proposed framework for compliance perception. It leverages tactile image sequences and jointly analyzes temporal contact forces and deformation features to estimate and classify the Shore-hardness of deformable objects.

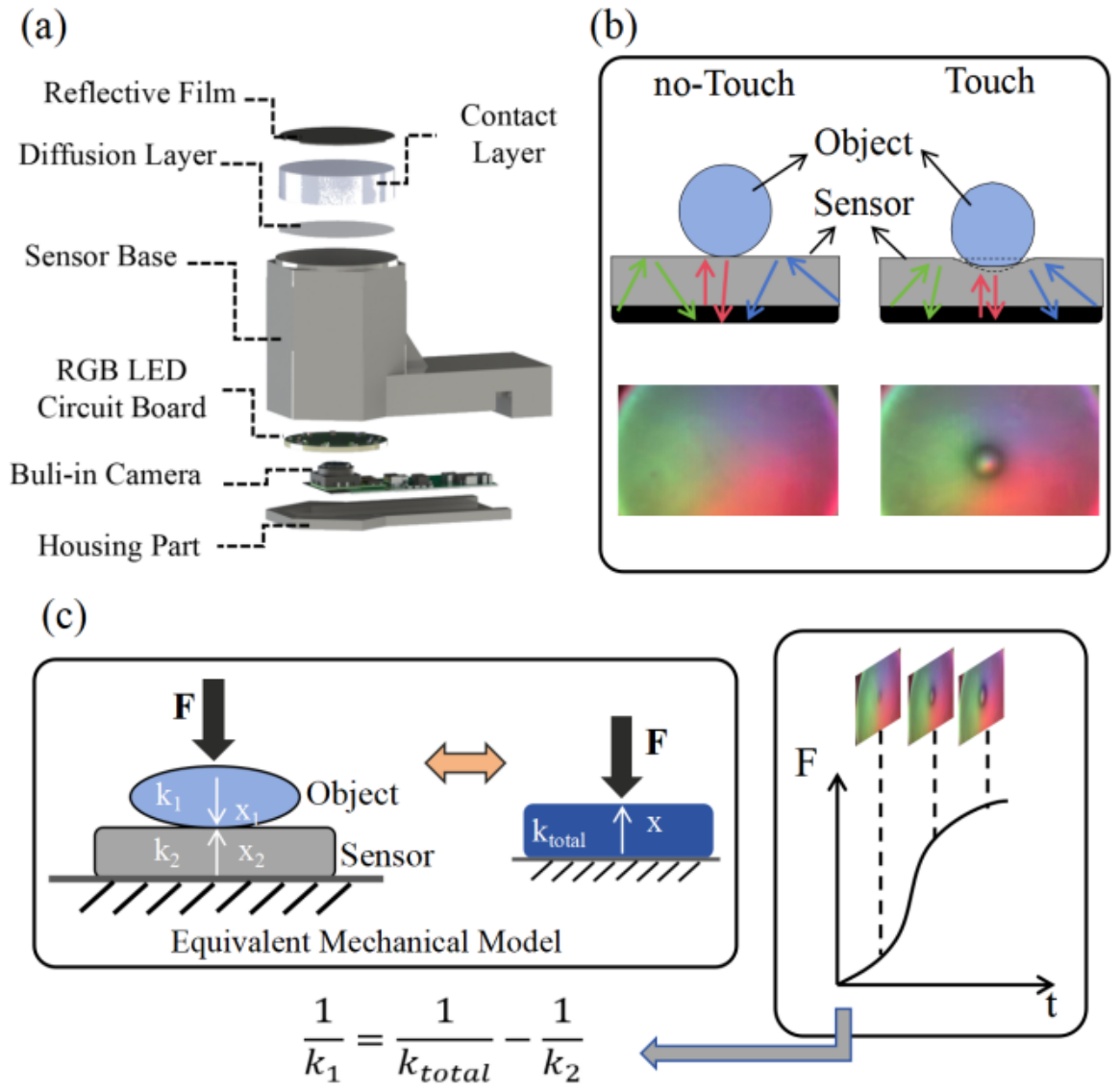

Fig. 2. Sensor structure and principle of compliance sensing. (a) Exploded view of sensor architecture. (b) Schematic of the sensor's response to contact deformation. (c) Equivalent mechanical model of object compliance under contact.

## II. Related Works

Vision-based tactile sensors leverage optical imaging to capture contact interface dynamics, with GelSight standing as the most representative system. These sensors analyze image variations induced by contact deformation to simultaneously resolve multiple physical quantities, including contact force [14-16], and geometric features [17-19]. Specifically, existing research often employs photometric stereo techniques to reconstruct high-resolution surface deformation fields [20-21], or utilizes deep learning models to establish a mapping between image and contact forces for force magnitude estimation [15-16]. Furthermore, by fusing this multi-modal sensory data with physical models, such systems can derive object physical properties, such as the object compliance [4-6]. Estimating object compliance by sensing the contact interface deformation field via a vision-based tactile sensor has emerged as an effective approach for categorizing soft objects. For instance, Fahmy et al. [22] captured a sequence of deformation field images during contact, achieving 92% accuracy in classifying avocado ripeness.

Similarly, other studies have leveraged deformation field perception combined with neural networks to identify ripeness based on compliance features [23-24]. However, these methods are predominantly qualitative, focusing on classification rather than precise quantitative compliance assessment. To address this limitation, traditional quantitative approaches typically apply Hooke's Law to gripping force and displacement measurements [25], or analyze temporal force dynamics to quantify object compliance [26-28]. For example, Su et al. [26] utilized distributed pressure signals from tactile sensors to achieve quantitative compliance estimation. Nevertheless, methods relying solely on force or macroscopic displacement often lack the rich spatial details of the contact interface's geometric deformation, limiting their ability to distinguish compliance in objects with complex geometries.

Recent studies have explored compliance estimation and multimodal tactile perception using vision-based or optical tactile sensors. Burgess et al. [4] estimated Young's modulus from single grasps, while Susini et al. [12] and Pagnanelli et al. [28] investigated compliance discrimination using biomimetic optical tactile sensing. Other works focused on hardness classification or stiffness-related recognition [5, 7, 10], and recent multimodal tactile sensors have demonstrated force, geometry, and contact-state perception capabilities [14-17]. Recent studies have also explored temporal feature extraction using Transformer-based models [22-24], as well as feature-extraction strategies based on photometric stereo, CNN visual features, or manually designed physical parameters. Unlike these methods, our framework uses experimentally measured Shore hardness as the reference label and fuses temporal force sequences with deformation-field features for Shore-hardness classification and compliant-object state recognition, rather than direct quantitative compliance estimation.

## III. Tactile Sensor Principles and Performance

To address the need for object compliance recognition, we developed a vision-based tactile sensor inspired by the GelSight sensing platform [9].

### A. Principles of Sensor Compliance Perception

The vision-based tactile sensor developed in this work builds on the classic GelSight architecture with iterative refinements, with its exploded structure shown in Fig. 2(a). The sensor employs a modular integration scheme: the topmost sensing layer comprises an aluminum reflective film and an elastomeric polydimethylsiloxane (PDMS) contact layer for tactile transduction; a Red-Green-Blue Light-Emitting Diode board acts as the illumination source, projecting uniform light onto the contact layer via a diffuser; a miniature embedded camera records the tactile image sequences; all components are assembled into a lightweight polylactic acid (PLA) housing.

The sensor operates via a vision-based tactile transduction mechanism that leverages the principle of light field variations. As shown in Fig. 2(b), when the sensor makes contact with a flexible sphere (6 mm diameter, 50 HA), the elastomeric contact layer and reflective film undergo concurrent deformation, which alters the path of reflected light within the elastomer. The embedded camera records the resulting variations in the spatial light field distribution, indirectly encoding the deformation and force information throughout the contact process. Recovering these two physical quantities forms the foundation for achieving object compliance perception.

As shown in Fig. 2(c), the force and deformation characteristics captured by the sensor during contact stem from the coupled interaction between the target object and the sensor's elastomeric layer. The entire contact system behaves equivalently to a mechanical model describing the target object's compressive deformation against a rigid plane.

Accordingly, we analyzed this process to theoretically verify the feasibility of leveraging this coupled interaction to enable the sensor to perceive the target object's compliance. The contact system is simplified and modeled as a series spring configuration [29] for analytical tractability. Assume that the system’s external force input is $F$; the force-induced deformation of the contact object and sensor elastomer satisfies:

$$F = k_1 \cdot x_1 = k_2 \cdot x_2 \tag{1}$$

Herein, $k_1$ denotes the stiffness of the contact object, with $x_1$ representing the compression of its contact area; $k_2$ is the stiffness of the sensor’s contact surface, and $x_2$ stands for the compression of the elastomer’s contact area.

If the contact object and the system's elastomer are equivalent to an integral with a stiffness of $k_{total}$, the total deformation $x_{total}$ under the external force $F$ satisfies:

$$F = k_{total} \cdot x_{total} \tag{2}$$

Combined with the extended series spring total stiffness formula derived from Hooke's Law

$$k_{total} = \left(\sum_{i=1}^{n} \frac{1}{k_i}\right)^{-1} \tag{3}$$

The contact process can be conceptually represented by a simplified series-contact model, in which the target object and the sensor elastomer jointly contribute to the total deformation under external loading. This model is used only to illustrate the coupled deformation relationship between the object and the tactile sensor, rather than to quantitatively calculate the stiffness or Shore hardness. It should be noted that the PDMS contact layer exhibits viscoelastic behavior, and its effective mechanical response may depend on the indentation rate, loading history, and contact geometry. Therefore, the sensor elastomer cannot be simply treated as a constant stiffness element under dynamic indentation. In this work, the Shore hardness values of the samples were directly measured using a Shore durometer and used as reference labels. The proposed perception pipeline does not derive hardness from the simplified mechanical model. Instead, the temporal force sequence inferred from tactile images is used as a dynamic force-response feature and is fused with deformation-field information for compliance perception.

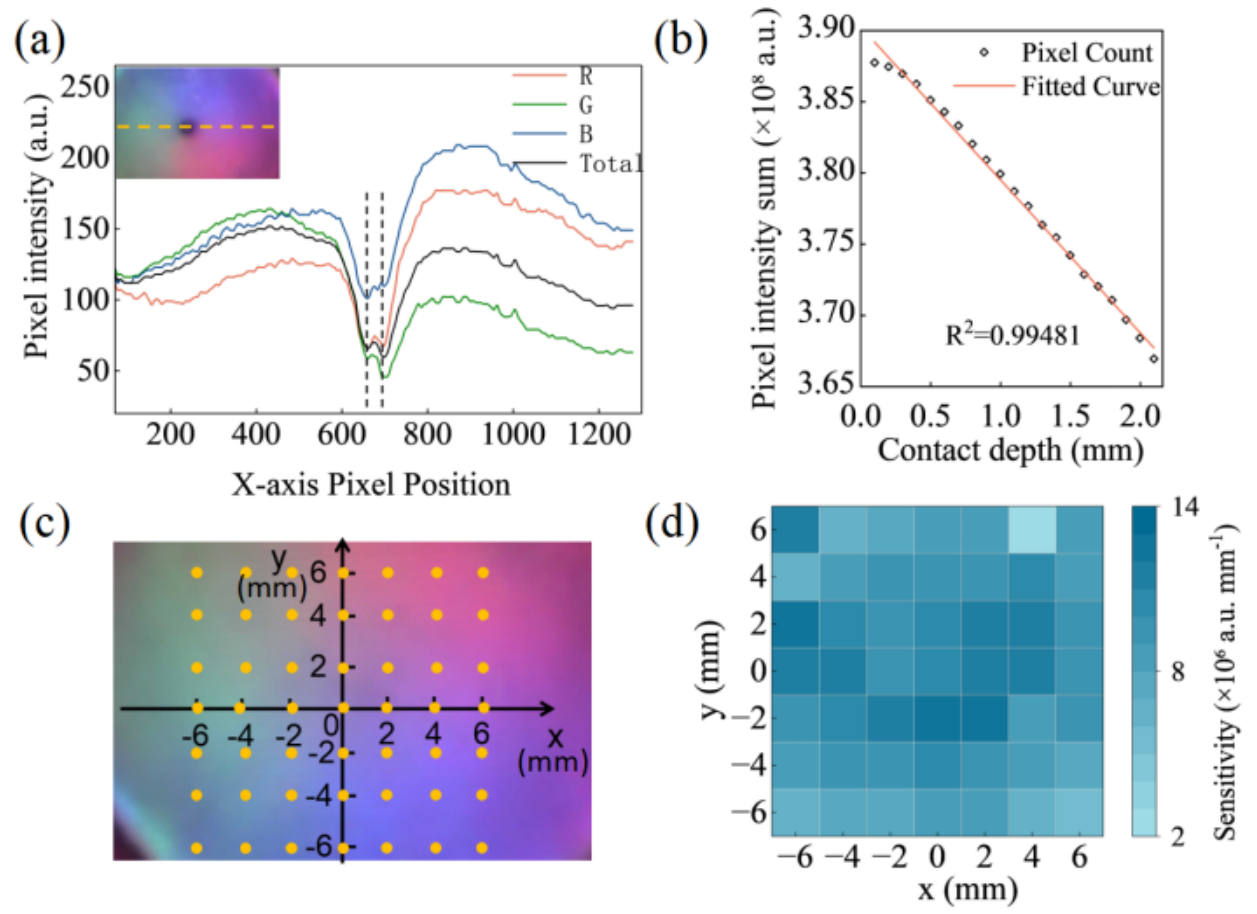

Fig. 3. Sensor performance characterization. (a) Two-point discrimination threshold curve. (b) Tactile image brightness versus indentation depth. (c) Spatial distribution of contact points across the sensing area. (d) Sensitivity distribution across different contact locations.

### *B. Basic Performance Test of the Sensor*

Accurately perceiving object compliance with this sensor hinges on the precise measurement of contact force and deformation. Specifically, force perception is related to changes in the tactile image captured during contact, while deformation perception depends on the sensor's spatial response capability. Therefore, we characterized the sensor's performance from two key dimensions: spatial resolution and sensitivity.

The two-point discrimination threshold, a standard metric for tactile spatial resolution, was measured using rigid dual-tip probes with spacings from 0.1 mm to 0.8 mm in 0.1 mm increments. Ten trials were conducted for each spacing at a constant indentation depth of 2 mm, and the pixel-brightness distribution along the Y-axis of the contact region was analyzed. As shown in Fig. 3(a), two distinct brightness-reduction peaks appeared when the probe spacing reached 0.3 mm, corresponding to the two contact points. The measured peak interval was 41 pixels, approximately 0.309 mm, which agreed well with the physical spacing. For spacings ≤0.2 mm, the two peaks merged into a single feature, making the contacts indistinguishable. Therefore, the sensor achieved a two-point discrimination threshold of approximately 0.3 mm, demonstrating its ability to resolve sub-millimeter deformation features. The sensor sensitivity was evaluated by analyzing the relationship between indentation depth and pixel brightness. A rigid cylindrical indenter with a diameter of 1 mm was progressively pressed into the sensor from 0.1 mm to 1.6 mm in 0.1 mm increments, while tactile images were synchronously acquired. As the indentation depth increased, the contact region gradually expanded, and the RGB channel intensities decreased linearly. Linear regression of pixel brightness versus indentation depth yielded a coefficient of determination of ($R^2$ = 0.99481), as shown in Fig. 3(b), confirming a strong linear relationship. This result provides a reliable basis for inferring contact force from image signals in subsequent compliance recognition.

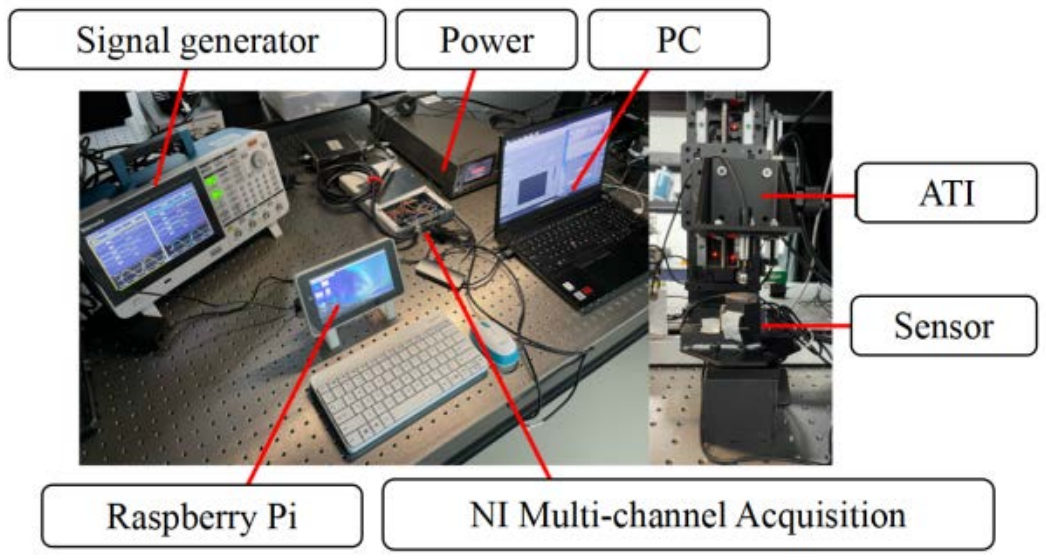

Fig. 4. Equipment setup for calibrating the relationship between force and tactile images.

Due to the inherent non-uniformity of light field distribution, sensitivity may vary across different sensor regions. To characterize this spatial variation, a 7×7 grid sampling strategy was employed, selecting 49 contact points with 2 mm intervals (Fig. 3(c)). As shown in Fig. 3(d), the sensor exhibits sensitivities of the same order of magnitude across the entire contact plane, demonstrating excellent spatial consistency. This uniformity ensures that a single calibration model can be applied globally without position-dependent compensation, thereby enabling stable and reliable compliance perception across the whole sensing interface.

### *C. Compliance Recognition Performance*

This study focuses on the temporal variation and deformation characteristics of forces to perceive the flexibility of objects. First, the sensor is enabled to determine the magnitude of the contact force, and then this force response is used to perceive compliance. The force-tactile image calibration device used in the experiment consists of a dynamometer (ATI NANO 17), a Raspberry Pi, and a displacement platform. The sensor is fixed on a horizontal platform to control the contact position, and the dynamometer is installed on a vertical displacement platform. By manipulating the force gauge, contact between the probe and the sensor can be established, allowing the construction of a tactile image dataset labeled with contact force. The specific setup is shown in Fig. 4.

In the compliance-recognition experiments, indentation was position-controlled at 0.75 mm/s, with a maximum depth of 1.5 mm, a 1 s holding period, and a 5 s cycle duration. Under this condition, fixed time intervals during loading corresponded to fixed indentation-depth increments. Over 4600 synchronized tactile image-force datasets were collected from samples with different hardness values and contact depths. A ResNet-50-based network was trained to model the tactile image-force relationship, achieving an absolute force prediction error of 0.06 N over the 0-12 N range and a minimum force resolution of 0.091 N. These results enable reliable temporal force extraction for subsequent compliance recognition.

Furthermore, building on the force information extracted from the tactile images, we capture the full contact process and decouple the force signals to investigate the differences in the temporal evolution of contact force across objects with varying hardness. First, we calibrated eight standard elastomer samples using a Shore durometer, with their Shore hardness ranging from 10 HA to 80 HA. The deviation between the measured and nominal values was less than 1%, verifying the reliability of the experiments. We then indented the sensor into each sample with a fixed indentation depth of 1.5 mm, and recorded the corresponding temporal evolution of contact force (Fig. 5(a)). The results show that within the same time window, harder samples generated larger contact forces. Moreover, the temporal force sequence was significantly correlated with material hardness, confirming that the sensor can effectively perceive the compliance of soft objects with varying different Shore-hardness levels.

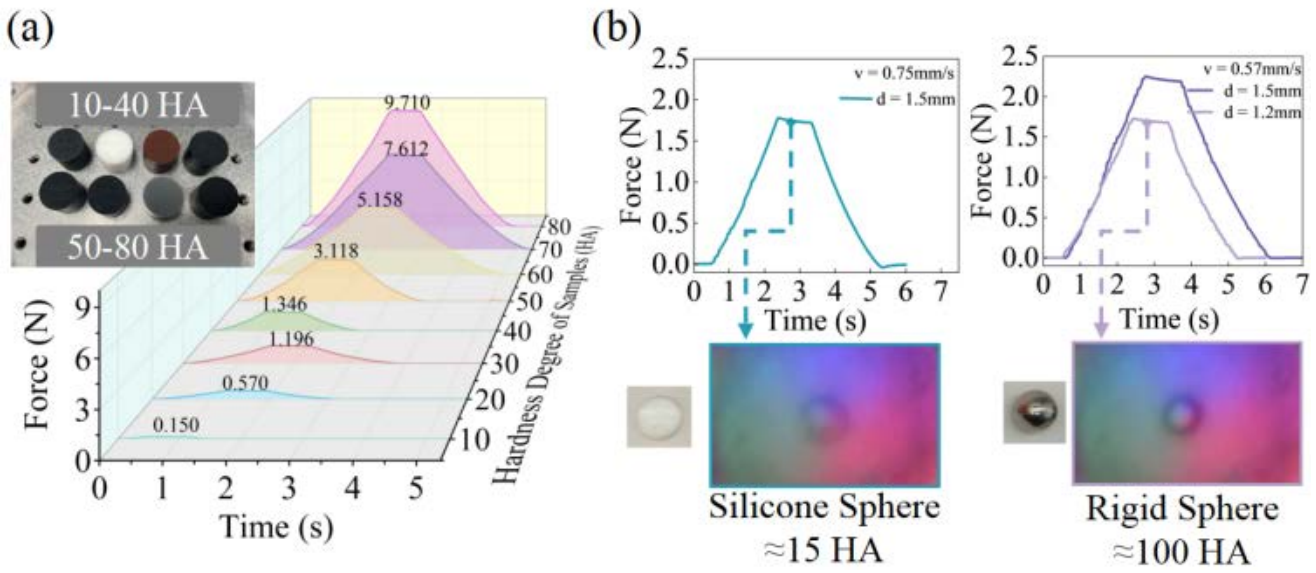

Fig. 5. Force responses and deformation fields of objects with different hardness. (a)Force response curves of objects with different hardness during dynamic contact. (b) Comparison between a silicone sphere (≈15 HA) and a rigid sphere (≈100 HA) with similar geometry.

To clarify the influence of loading conditions on temporal force responses, we tested two spherical samples with the same shape, size (6 mm diameter), and curvature but different hardness values: a silicone sphere (≈15 HA) and a rigid sphere (≈100 HA), as shown in Fig. 5(b). At the same maximum indentation depth of 1.5 mm, the silicone sphere indented at 0.75 mm/s and the rigid sphere indented at 0.57 mm/s showed similar force-rising gradients, but different peak forces due to different loading durations (1.75 N and 2.30 N, respectively). When the indentation depth of the rigid sphere was reduced from 1.5 mm to 1.2 mm, its peak force and force-evolution trend became close to those of the silicone sphere, making force-only discrimination difficult. In contrast, the deformation fields remained distinct: the silicone sphere showed a smoother boundary transition and larger affected region, whereas the rigid sphere showed a sharper boundary and more localized deformation. These results indicate that temporal force responses are affected by indentation velocity and depth, highlighting the need to fuse them with deformation-field information for robust compliance perception.

## IV. Experiments and Results

### *A. Cross-Modal Influence Analysis*

The goal of this work is to enable compliance perception of deformable objects during contact using the proposed tactile sensor, which relies on sensing the deformation features and temporal evolution of contact force. However, it remains to be verified whether the sensor system can effectively separate and accurately sense these two distinct sources of information, while avoiding cross-modal interference. To this end, we designed a controlled experiment (Fig. 6(a)) to validate this. Specifically, this experiment focuses on the problem of modal isolation, testing the independent processing and perception performance of object compliance sensing and deformation field estimation, respectively. As shown in Fig. 6(b), 4 representative geometric shapes were selected: circle, square, triangle, and T-shape. For each shape, 8 hardness grade samples (10–80 HA) were set to ensure representativeness in the hardness dimension.

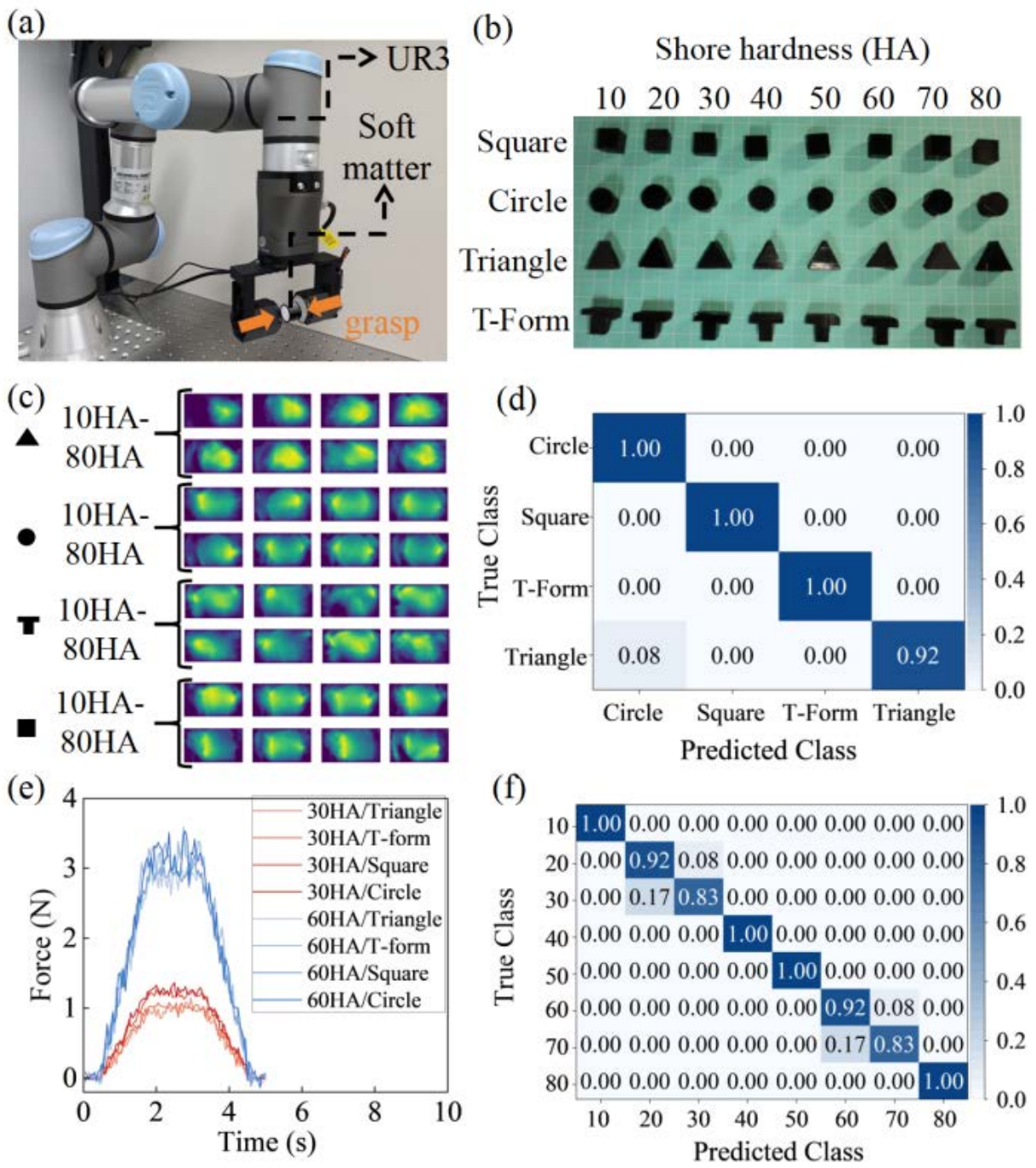

Fig. 6. Cross-modal influence evaluation. (a)Experimental testing system. (b) Experimental samples of different shapes and hardness levels. (c)Depth map information of samples with the same shape but different hardness. (d) Shape recognition confusion matrix. (e) Force response curves of samples with the same hardness but different shapes. (f)Hardness recognition confusion matrix.

The experiment collected contact images of objects with four different shapes and hardness levels at a contact depth of 1mm and generated depth maps (Fig. 6(c)) to explore the effectiveness of the system framework in recognizing the geometric feature. Quantitative evaluation experiments were designed to verify and quantify the sensing system's Three-Dimensional reconstruction capability. It can be seen that although the objects vary in hardness, the core geometric information is still retained, verifying the validity of depth maps as a basis for geometric differentiation. Furthermore, contact images of objects with the same shape and hardness properties were collected at depths ranging from 0.2 mm to 1.0 mm in 0.2 mm steps. We employed an image classification algorithm based on a convolutional neural network (CNN). The resulting shape recognition confusion matrix (Fig. 6(d)) shows an overall accuracy of 98.0%, demonstrating that the proposed perception system possesses deformation recognition capabilities even in objects of varying hardness.

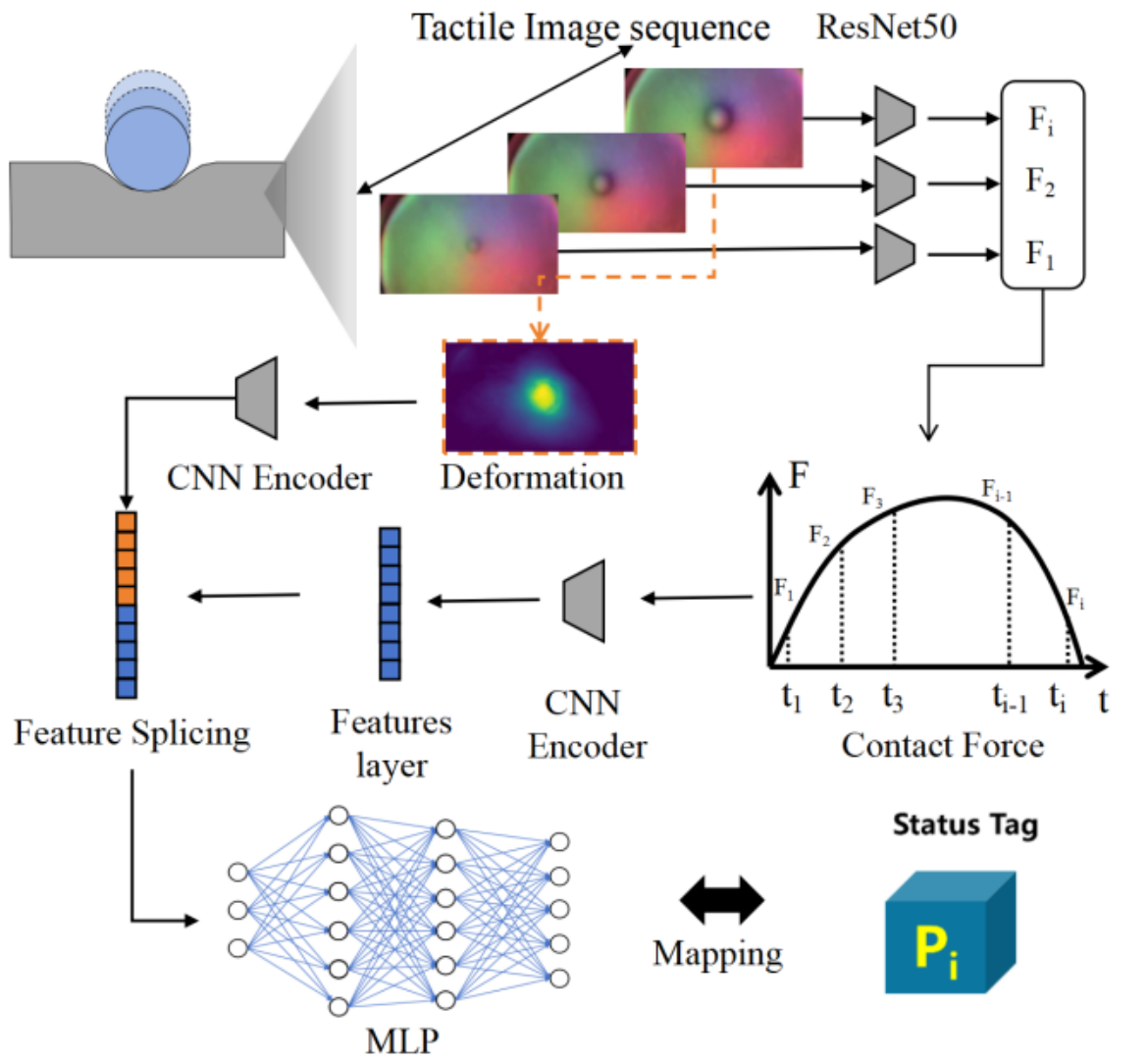

Fig. 7. The proposed architecture for object compliance perception and estimation. Temporal force sequences and deformation field depth information are fed into a neural network model to output the final object compliance classification results.

To investigate the influence of geometric shapes on compliance recognition, force response curves of objects with identical hardness but different shapes were collected. Data for Rockwell hardness 30 HA and 60 HA (Fig. 6(e)) show that force curves differ between hardness levels but are essentially consistent for the same hardness across shapes (including force temporal gradient and maximum value). Despite shape variations, force response curves remain closely related to hardness, indicating the method accurately captures hardness mechanical information across geometries. When the fusion framework recognized four objects of different shapes but the same hardness, the output hardness recognition confusion matrix (Fig. 6(f)) showed an accuracy of 93.75%. This indicates geometric features minimally affect hardness recognition under the proposed system, confirming the two-stream network's strong ability to recognize hardness characteristics.

Overall, in terms of the reliability of individual modalities, depth maps can stably identify the deformation features of deformable objects, while temporal force sequences can accurately distinguish compliance-related features. These results jointly lay the foundation for realizing object compliance perception by leveraging both temporal force sequences and deformation field sensing.

### B. Perception Algorithm

To further validate the compliance perception capability of our proposed framework, this section focuses on evaluating its performance in real-world environments. As shown in Fig. 7, an object compliance perception strategy is proposed, which fuses temporal force sequences and deformation field information. Specifically, an image-force correlation model is first constructed to infer the dynamic variations of contact force, generating temporal force sequences to capture the mechanical response of deformable materials. $F_1$, $F_2$, and $F_3$ denote representative force samples selected at fixed indentation-depth increments during the loading stage. Since the indentation velocity was fixed at 0.75 mm/s, these time-interval samples are equivalent to fixed depth-based samples.

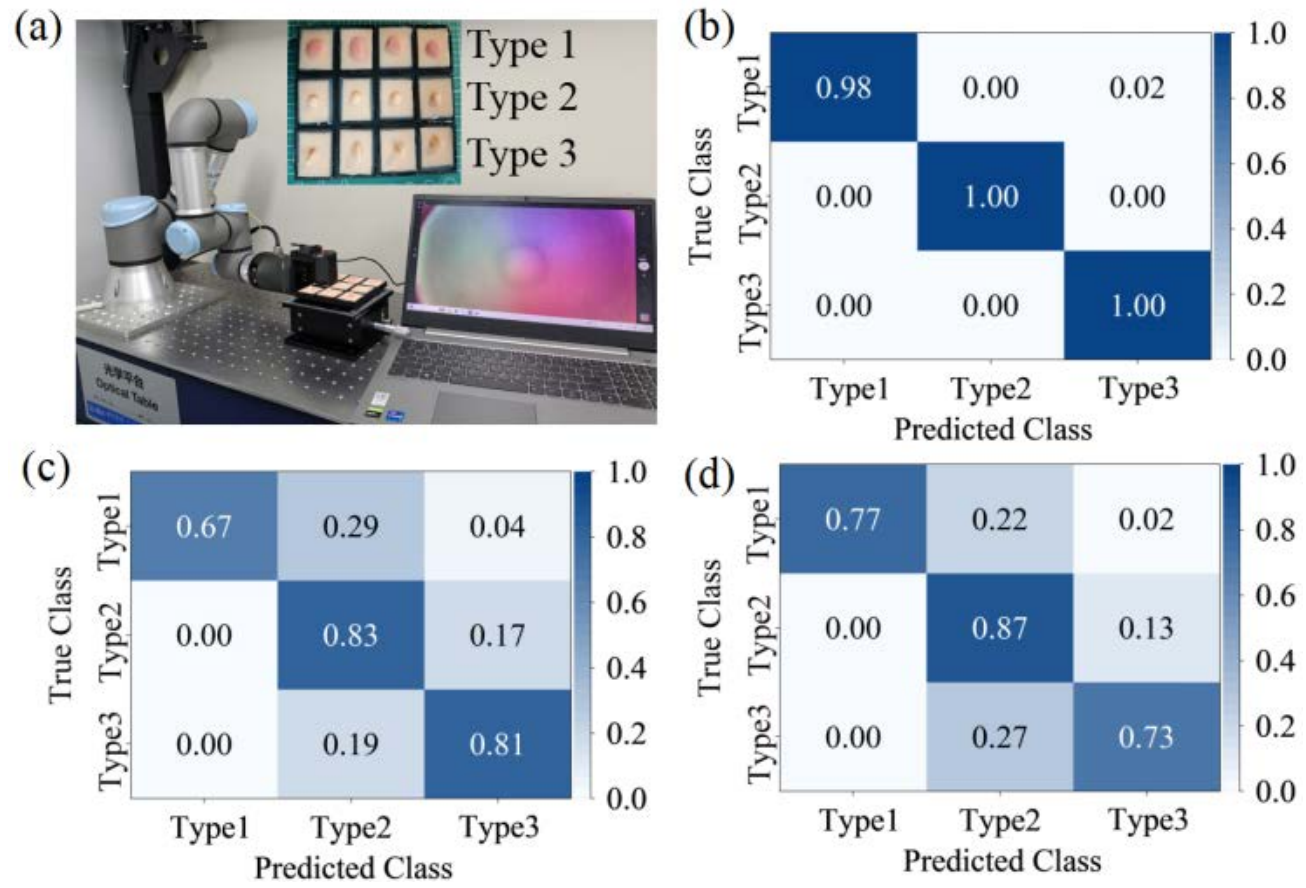

Fig. 8. Compliant object recognition based on perception. (a) Dual-finger manipulator system. (b) Confusion matrix of object classification effect based on dual-modal perception. (c)Simulated compliant objects classification effect based on perception of internal hardness features. (d)Simulated compliant objects classification effect based on perception of external geometric features.

In addition, photometric stereo was used to extract object shape and surface-structure information from tactile image sequences, generating depth maps as deformation-field representations. The deformation-field features were extracted using ResNet-18, while the temporal force-sequence features were extracted using a 1D-CNN. These two feature vectors were then concatenated and fed into an MLP for multi-modal fusion and compliance classification, enabling joint recognition of object geometry and mechanical properties. More advanced architectures, such as attention-based multi-modal fusion, Transformer-based temporal models, stronger CNN backbones, and alternative feature extractors, may be further explored in future work.

For the compliance-recognition task, the dataset was divided into training, validation, and test sets with proportions of 70%, 15%, and 15%, respectively. To avoid data leakage, the classification dataset was split into training and test sets at the level of independent indentation video clips rather than individual image frames. All image frames and the corresponding temporal force sequence from the same video clip were assigned to the same subset, while different clips of the same object were collected from independent indentation trials with slight variations in contact position and orientation.

### *C. Integrated Application Validation*

To verify the proposed compliance perception strategy, an object perception platform was built, as shown in Fig. 8(a). Ecoflex-based flexible simulants with different hardness and surface morphologies were used to represent diverse compliant-object states. Specifically, Type 1 denotes soft samples with irregular boundaries, Type 2 denotes moderately hard samples with near-circular shapes, and Type 3 denotes hard strip-like samples with dense structures. Multiple sets of tactile data were collected using the robotic arm system and labeled according to hardness and morphology. The proposed fusion perception model was then used for joint morphology and hardness classification. As shown in Fig. 8(b), the system achieved an accuracy of 99.3% through simulated palpation, demonstrating its ability to distinguish compliant samples with different physical characteristics.

**TABLE I**

OBJECTS OF VARYING SIZES AND HARDNESS MODELED TYPICAL BIOLOGICAL TISSUES AND STRUCTURES.

| Sample ID | Diameter （mm） | Hardness (HA) | Typical biological structures |
|---|---|---|---|
| S1 | 4 | 38 | Articular cartilage; Meniscus; Tendon |
| S2 | 8 | 45.5 | Cancellous bone ; Cortical bone; Dentine |
| S3 | 8 | 57.5 | Compact bone ; Tooth root; Cervical bone |
| S4 | 17 | 66.5 | Cortical bone fragment; Tooth fragment; Osteoma |
| S5 | 8 | 100 | Enamel; Dense bone with pathological calcification; |

To systematically evaluate the contribution of each component, an ablation study was conducted with three settings: temporal-force-only, deformation-field-only, and dual-modal fusion. The temporal-force-only model used only the extracted force sequences as input and achieved a classification accuracy of 77.0% (Fig. 8(c)). The deformation-field-only model used only depth images and achieved an accuracy of 79.0% (Fig. 8(d)). In contrast, the proposed dual-modal fusion framework, which combines temporal force features with deformation-field features, achieved an accuracy of 99.3%. These results indicate that temporal force evolution and deformation-field information provide complementary cues for compliance recognition, and that their fusion is essential for robust classification.

The sensor was integrated with a UR robotic arm to construct a robotic dexterous-hand operating system, as shown in Fig. 9(a). For physical-property assessment, tactile perception is required to distinguish object hardness and shape. In this experiment, spheres with different sizes and hardness values were used as compliant-object samples. The five samples (S1–S5) exhibited distinct physical properties and were selected to provide references for applications involving biological tissue-mimetic materials. As shown in Table 1 [30], S1 represents soft materials analogous to articular cartilage; S2 and S3 correspond to harder materials similar to cancellous and cortical bone; S4 mimics tooth fragments or pathological bone hyperplasia; and S5 represents extremely hard materials such as enamel or pathologically calcified tissues.

The system performs object feature recognition using vision-based tactile sensors integrated into a robotic arm gripper, which maintains a 2.0 mm contact threshold. Contacting five samples, the system successfully decoupled surface morphology. Dynamic contact video streams were extracted, and the time-series variation of contact force was analyzed. Based on compliance and surface morphology, 625 experimental datasets were labeled and classified using the proposed fusion perception neural network model. The classification results (Fig. 9(b)) demonstrate the system's ability to accurately distinguish samples with varying surface morphologies and hardness, achieving an accuracy of 99.0%.

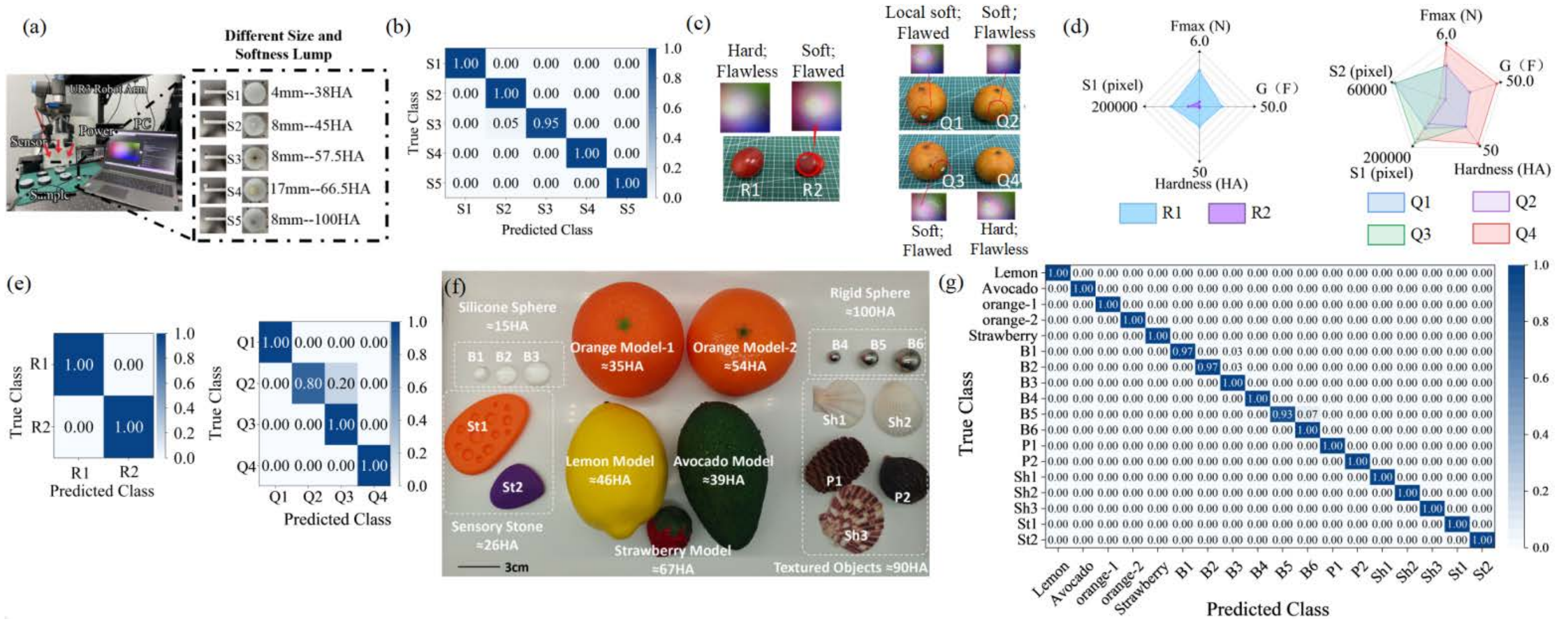

Fig. 9. Robotic integration and soft-object recognition. (a) Sensor mounted on the robotic dexterous hand and five test samples. (b) Confusion matrix of the fusion perception framework. (c) Tactile images of oranges and cherry tomatoes. (d) Extracted feature parameters. (e) Classification results of orange and cherry tomato states. (f) Extended 18-object set with variations in size, local curvature, surface texture, and Shore hardness. (g) Confusion matrix of the 18-class recognition experiment.

To evaluate the necessity of the proposed network framework, a control experiment was conducted by replacing CNN-based feature extraction with manually extracted features. Specifically, sphere radius derived from depth maps and measured Shore hardness were fed into an MLP for classification. This manual feature-based method achieved an accuracy of 98.5%, but still performed lower than the proposed fusion perception framework. The result indicates that limited manually designed features, such as radius and hardness, may fail to capture subtle deformation and force-profile differences, especially for objects with complex geometries or mechanical responses.

The experiment selected fruits with abnormal morphology to further verify the effectiveness of the sensing system in identifying daily physical objects. These fruits were grabbed and perceived by the robotic arm dexterous hand system integrated with a vision-based tactile sensor, as shown in Fig. 9(c). Among them, oranges were set in 4 states, including different defect areas and different local hardness, all of which are significantly distinguishable from normal fruits; cherry tomatoes were divided into two categories: hard ones without defects and softened ones with defects.

To validate the fusion framework's recognition effectiveness, preliminary feature extraction ensured distinct experimental object parameters. Fig. 9(d) details geometry and hardness analysis. For oranges, S1 (total contact texture area) and S2 (defective region) were analyzed alongside G(F) (force response/pressure gradient) and Hardness (actual hardness). Results confirm a positive correlation between orange compliance and G(F). Softer oranges correspond to lower G(F) values, and harder ones to higher values. Defective oranges were distinguishable by projected area; a radar chart computed defect area *via* pixel statistics, convertible to real size, confirming differential physical characteristics. For abnormal cherry tomatoes, optically resolution-limited surface defects precluded direct defect area measurement, necessitating the use of total contact texture area S2.

The experiment classified oranges (Q1-Q4) and cherry tomatoes (R1-R2) into distinct quality grades. As shown in Fig. 9(e), the system achieved a classification and recognition accuracy of no less than 98.5% based on the objects' physical parameter differences, validating the effectiveness of vision-based tactile sensors in robotic grasping and perception tasks. In practical scenarios like fruit grasping, the system can identify target objects and extract modality-specific mechanical and deformation field features based on their physical properties.

To further evaluate the robustness of the proposed method under more complex geometries, we additionally tested 18 objects with different local curvatures, sizes, surface textures, and Shore hardness levels, as shown in Fig. 9(f). These samples included silicone spheres (B1-B3), rigid spheres(B4-B6), fruit-shaped compliant models, sensory stones (St1, St2), and textured objects (Sh1-Sh3, P1, P2). Unlike the controlled planar-outline samples, these objects introduced more realistic variations in curvature and contact geometry. As shown in Fig. 9(g), the proposed fusion framework achieved an overall recognition accuracy of approximately 99.3%, demonstrating its ability to distinguish compliant objects under more diverse geometric and mechanical conditions. This fully demonstrates the necessity of the proposed object compliance perception strategy that fuses temporal force sequences and deformation field information. (See videos S1).

## V. Conclusion

This paper presents a vision-based tactile perception framework for object compliance recognition by integrating temporal force sequences with deformation-field information. The proposed method addresses the limitations of force-only and deformation-only approaches by jointly modeling dynamic mechanical responses and contact deformation features. Experiments on Shore-hardness classification, compliant-object recognition, and abnormal fruit detection demonstrate that the proposed fusion strategy improves recognition accuracy and robustness under different material and geometric conditions, achieving an overall recognition accuracy exceeding 98.5% in practical application scenarios. These results indicate that combining force evolution and deformation-field cues is effective for interactive tactile perception and provides a useful basis for robotic applications involving soft-object recognition and physical-property assessment. The compliance recognized in this work represents the overall response of the contacted region under controlled indentation, rather than a complete spatial stiffness distribution. The proposed fusion framework shows robustness for objects with similar compliance-related characteristics and provides a basis for improving unseen-object generalization through larger datasets, stricter object-level splits, and further algorithm optimization. Future work will further improve both the sensor design and processing algorithms to obtain distributed deformation/force-field information, enabling local stiffness mapping for objects with spatially non-uniform mechanical properties.